\documentclass[10pt,twocolumn,letterpaper]{article}

\usepackage{iccv}
\usepackage{times}
\usepackage{epsfig}
\usepackage{graphicx}
\usepackage{amsmath}
\usepackage{amssymb}

\usepackage{booktabs}
\usepackage{multirow}

\usepackage[breaklinks=true,bookmarks=false]{hyperref}

\iccvfinalcopy % *** Uncomment this line for the final submission

\def\iccvPaperID{3761} % *** Enter the ICCV Paper ID here

\ificcvfinal\fi

\begin{document}

%%%%%%%%% TITLE
\title{HBONet: Harmonious Bottleneck on Two Orthogonal Dimensions}

\author{
Duo Li\thanks{Equal contribution. This work was done when Duo Li and Aojun Zhou were interns at Intel Labs China, supervised by Anbang Yao who is responsible for correspondence. Intern Duo Li performed most experiments.} \quad Aojun Zhou$^{*}$ \quad Anbang Yao \\
Intel Labs China\\
{\tt\small \{duo.li, aojun.zhou, anbang.yao\}@intel.com}
% Intern Duo Li conducted experimental validation.
%First Author\\
%\and
}

\maketitle
\ificcvfinal\thispagestyle{empty}\fi

%%%%%%%%% ABSTRACT
\begin{abstract}
	MobileNets, a class of top-performing convolutional neural network architectures in terms of accuracy and efficiency trade-off, are increasingly used in many resource-aware vision applications. In this paper, we present Harmonious Bottleneck on two Orthogonal dimensions (HBO), a novel architecture unit, specially tailored to boost the accuracy of extremely lightweight MobileNets at the level of less than 40 MFLOPs. Unlike existing bottleneck designs that mainly focus on exploring the interdependencies among the channels of either groupwise or depthwise convolutional features, our HBO improves bottleneck representation while maintaining similar complexity via jointly encoding the feature interdependencies across both spatial and channel dimensions. It has two reciprocal components, namely spatial contraction-expansion and channel expansion-contraction, nested in a bilaterally symmetric structure. The combination of two interdependent transformations performing on orthogonal dimensions of feature maps enhances the representation and generalization ability of our proposed module, guaranteeing compelling performance with limited computational resource and power. By replacing the original bottlenecks in MobileNetV2 backbone with HBO modules, we construct HBONets which are evaluated on ImageNet classification, PASCAL VOC object detection and Market-1501 person re-identification. Extensive experiments show that with the severe constraint of computational budget our models outperform MobileNetV2 counterparts by remarkable margins of at most 6.6\%, 6.3\% and 5.0\% on the above benchmarks respectively. Code and pretrained models are available at \url{https://github.com/d-li14/HBONet}.
\end{abstract}

%%%%%%%%% BODY TEXT

\section{Introduction}
By winning ImageNet classification challenge 2012 with a large margin, AlexNet~\cite{ref01} ignited the surge of deep Convolutional Neural Networks (CNNs) in a variety of computer vision tasks such as image classification~\cite{ref02}, object detection~\cite{ref03} and semantic segmentation~\cite{ref04}. In order to achieve higher accuracy, there shows an evident trend to make CNN architectures deeper and topological connections more sophisticated in recent literature~\cite{ref05,ref06,ref07,ref08,ref09,ref10}. However, top-performing CNNs usually come with tremendous storage consumption and heavy computational cost, which prohibits the feasibility of their practical deployment in resource-constrained environments.

\begin{figure}
	\centering
	\vspace{-2.0em}
	\includegraphics[width=\linewidth]{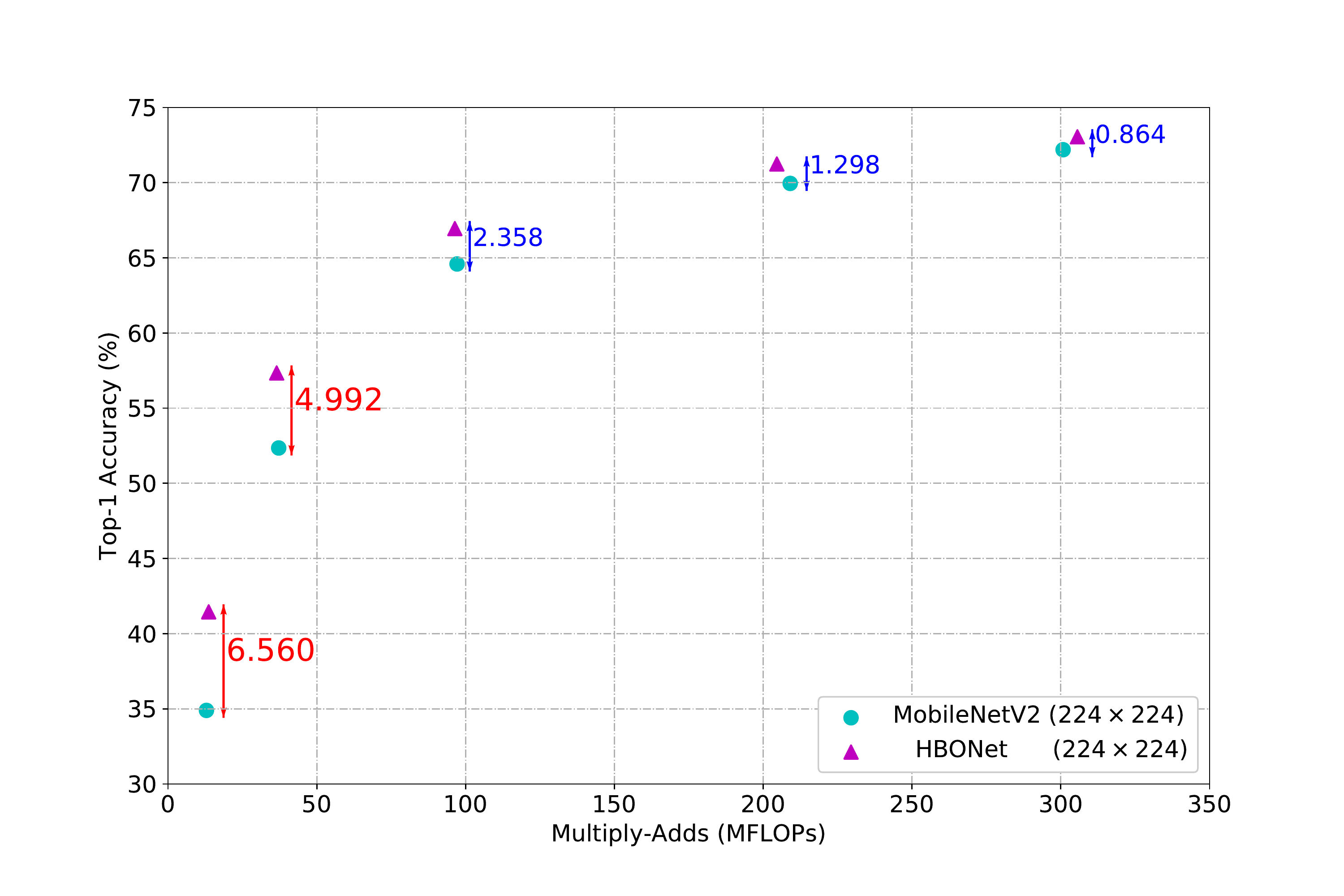}
	\vspace{-2.0em}
	\caption{
		Comparison of top-1 accuracy and FLOPs for MobileNetV2 and our HBONet models with five different width multipliers, tested on the ImageNet validation set with the single crop. Our models demonstrate increasingly large margins against MobileNetV2 counterparts when the computational budget goes to less than 40 MFLOPs.
	}
	\label{fig:flops-accuracy}
	\vspace{-1.5em}
\end{figure}

Towards the problem above, numerous research efforts have been devoted to engineering lightweight CNN architectures from scratch with expertise. Among existing designs, the family of CNNs~\cite{Chollet_2017_CVPR,ref13,ref14,ref15,ref16,ref17,ref18,ref19} built upon depthwise separable convolutions is becoming the mainstream due to its leading performance in balancing accuracy and efficiency. A standard depthwise separable convolution, which consists of a depthwise convolution and a pointwise convolution, was originally presented in~\cite{ref20}. Later, it was extensively used in the Xception architecture~\cite{Chollet_2017_CVPR}. MobileNetV1~\cite{ref13}, the pioneering lightweight CNN backbone specially designed for vision applications on mobile and embedded devices, is mainly built on depthwise separable convolutional layers stacked in a straightforward way. ShuffleNetV1~\cite{ref14} uses residual bottlenecks harnessing pointwise group convolutions to reduce the complexity of 1x1 pointwise convolutions, and channel shuffle operations to enhance inter-channel correlations. Preserving the effective shuffling operation in ShuffleNetV1~\cite{ref14}, ShuffleNetV2~\cite{ref15} presents more hardware-aware modular designs in which the specific configuration of feature channels and the order of basic operations are adjusted to better match the proposed practical guidelines. MobileNetV2~\cite{ref16}, an advanced variant of MobileNetV1~\cite{ref13}, is based on an inverted residual structure with linear bottlenecks. Owing to improved information flow in the representation space, MobileNetV2 achieves much better trade-off between accuracy and efficiency compared with its predecessor. Following a similar design principle, each of these state-of-the-art lightweight architectures provides a spectrum of models at different computational complexities. However, they all perform unsatisfactorily at the level of less than 40 MFLOPs which is a necessary requirement in many extremely low-power platforms~\cite{ref15}. In this paper, we attempt to bridge this accuracy gap with a novel bottleneck design, taking MobileNetV2 backbone as a reference case without loss of generality.

Note that modular designs of MobileNets~\cite{ref13,ref16} and ShuffleNets~\cite{ref14,ref15} put an emphasis on the transformation from the perspective of feature channels, while neglecting to explore the orthogonal space of spatial feature scale. It provides further potential to shrink computational cost while retaining the comparable accuracy since feature map size is another principal element involved in the formula of complexity computation. In turn, there exists a chance to promote the performance given a certain amount of available computational resource. Motivated by this, we investigate both aspects and coordinate them in one novel bottleneck design called Harmonious Bottleneck on two Orthogonal dimensions (HBO), aiming to improve bottleneck representation ability from two complementary dimensions. In each HBO module, a spatial contraction operation is responsible to reduce input feature maps to a smaller size temporally, offering a capacity guarantee to increase computational efficiency. The following channel expansion-contraction component compensates for resulting side effect by encouraging informative features. Finally, a spatial expansion operation is performed to make output features have the same size as that of the output from the shortcut connection.

Summarily we make the following contributions to efficient yet accurate neural network architecture design:
\begin{itemize}
	\vspace{-2mm}
	\setlength\itemsep{0em}
	\item We present a bottleneck design named HBO, which subtly arranges spatial and channel transformations in a bilaterally symmetric layout for their mutual promotion. We notice that it has never been well studied in the research field of lightweight CNN design from both of these two orthogonal dimensions before us.
	\item We use HBOs to replace the original inverted bottlenecks in MobileNetV2 and construct HBONets. Benefiting from the conjugation of spatial and channel transformations, the performance of HBONet backbones exceeds that of MobileNetV2 counterparts by at most 6.6\%, 6.3\% and 5.0\% on different tasks and benchmarks under limited computational budgets, \eg less than 40 MFLOPs. To the best of our knowledge, we are the first to push the lower boundary with respect to the computational complexity of lightweight CNNs to such an extreme domain. Figure~\ref{fig:flops-accuracy} provides comprehensive comparative results under different computational budgets.
	\item Our proposed HBONet (1.0) surpasses state-of-the-art lightweight architectures on the challenging ImageNet benchmark at the level of 300 MFLOPs, achieving 73.1\% top-1 classification accuracy.
\end{itemize}

%-------------------------------------------------------------------------
\section{Related Work}
We summarize representative advances on efficient neural network architectures and transformations with respect to spatial feature dimensions as follows.

\textbf{Neural Network Compression.} By default, deep neural networks are trained with 32-bit floating-point parameters, thus network quantization is a natural way to obtain more efficient and smaller models using low-bit parameters. For instance, ~\cite{ref25} and ~\cite{ref26} adopt 16-bit and 8-bit fixed-point implementation respectively, and~\cite{ref27,ref28,zhou2017,ref29} attempt to train binary/ternary neural networks either from the pre-trained models or from scratch. Network pruning presents another promising way to convert dense neural network models into sparse equivalents without loss of predication accuracy. This line of research includes network parameter pruning~\cite{ref30}, filter pruning~\cite{ref31} and channel pruning~\cite{ref32}. Deep neural networks can also be compressed and accelerated via factorized networks~\cite{ref33,ref34} which utilize filter factorization techniques to reduce the computational cost of convolutional layers. However, our research efforts have been mainly invested in designing hand-engineered efficient neural network architectures from scratch, without cumbersome iterative training and fine-tuning in some compression methodology.

\textbf{Computational Efficient Neural Networks.} In order to reduce parameter size and computational burden, many top-performing CNNs adopt group convolution in which input channels are split into different groups and each convolution only operates on the corresponding channel group. As discussed in the last section, MobileNets~\cite{ref13,ref16} and ShuffleNets~\cite{ref14,ref15} heavily rely on depthwise convolution during their construction process which is an extreme and popular case of standard group convolution. Group convolution was first used in AlexNet~\cite{ref01} to make its training applicable on two separate GPUs. Inception series~\cite{ref11,ref12} customizes group convolution application by coupling it with multi-branch design. SqueezeNet~\cite{ref21} is based on a very small inception-like fire module and achieves AlexNet-level accuracy with 50$\times$ fewer parameters. ResNeXt~\cite{ref09} integrates group convolution into the residual block and improves efficiency through introducing a new dimension called ``cardinality''. CondenseNet~\cite{ref22} flexibly combines densely connected group convolutions with a filter pruning strategy to remove redundant connections. IGCNets~\cite{ref17,ref23,ref24} introduce two successive interleaved group convolutions to enhance feature representation ability. Recently, NAS~\cite{ref18} and ENAS~\cite{ref19} use reinforcement learning to automatically search an optimal neural network architecture based on a set of pre-defined operation units including depthwise convolution, opening up a new neural network design direction.

\textbf{Spatial Feature Scale.} Being primarily engineered for ImageNet classification task, prevalent CNN backbones including but not limited to AlexNet~\cite{ref01}, VGGNet~\cite{ref05}, GoogLeNet~\cite{ref06}, ResNet~\cite{ref07}, DenseNet~\cite{ref08}, ResNeXt~\cite{ref09}, SENet~\cite{ref10}, MobileNets~\cite{ref13,ref16} and ShuffleNets~\cite{ref14,ref15} follow a common design principle: the spatial feature scale of convolutional layers starts with a relatively large value (\eg 224$\times$224) and then reduces by a factor of 2 after each downsampling layer using either pooling operations or convolutions with stride 2, until reaching its desired value (\eg $1\times1$), no matter how deep the network is. This spatial feature downsampling design over the network body facilitates hierarchical feature extraction at different scales, meanwhile balancing layer-wise distribution of the computational cost of the whole network. However, as for the building block design regarding all these backbones, the spatial feature scale usually keeps unchanged across all layers inside one single block, except for certain layers responsible for downsampling located in the entry of few blocks.

In order to achieve pixelwise predication outputs from arbitrary-sized input, Long et al.~\cite{ref04} propose fully convolutional networks which combine multi-scale feature maps from shallow, intermediate and deep layers of a classification network via deconvolution operators for upsampling. U-Net~\cite{ref38} further develops this idea with a U-shaped architecture in which several expansion blocks are stacked for successive upsampling operations. Such kind of conv-deconv and encoder-decoder architectures are also used in other vision tasks such as style transfer~\cite{ref39} and image generation~\cite{ref40}. To address human pose estimation, ~\cite{ref41} presents repeated hourglass modules where each of them has a downsampling-upsampling symmetric structure. Recently, ~\cite{ref42} proposes a more simple spatial module design to replace any single convolutional layer and accelerate corresponding convolution operations. In our proposed HBO design, a channel expansion-contraction module is wrapped up in a spatial contraction-expansion component in the micro-architecture, resembling the conv-deconv or encoder-decoder framework in the macro-architecture.

%------------------------------------------------------------------------
\section{Proposed CNN Architecture}

In this section, we first describe our core bottleneck design HBO which delicately couples two structurally symmetric components: bottleneck in spatial dimension and inverted bottleneck in channel dimension. We then describe our HBO exemplars used to construct the HBONet architecture.

\subsection{Depthwise Separable Convolutions}
Modern CNNs tend to have no fully connected layer regardless of the last prediction layer with a softmax function, thus convolutional layers occupy most of the computational cost and parameters of the whole model. Depthwise separable convolution serves as a computational effective equivalent of standard convolution and is utilized as the most critical ingredient in many efficient CNN architectures~\cite{ref13,ref14,ref15,ref16,ref17,ref18,ref19}, which is extensively employed in our HBO design without exception. A standard convolutional layer directly transforms an $h\times w\times c_1$ input feature tensor into an $h\times w\times c_2$ output feature tensor by a $c_1\times k\times k\times c_2$ convolutional kernel, where $h\times w$, $c_1/c_2$ and $k\times k$ are the spatial size of input/output feature maps, the number of input/output feature channels and the convolutional kernel size, respectively. Neglecting the bias terms, it has the computational cost of $h\times w\times c_1\times c_2\times k\times k$. A depthwise separable convolutional layer decomposes standard convolution operation from one stage into two stages. It starts with a depthwise convolution that performs a $k\times k$ convolution on each channel of the input feature tensor, and follows with a $1 \times 1$ pointwise convolution to project the concatenated $c_1$ channels produced by the depthwise convolution to a new space with the desired channel size of $c_2$, introducing interactions among different channels as well. By performing convolutions in this way, a depthwise separable convolutional layer only has the computational cost of
\begin{equation}\label{eq:01}
\begin{aligned}
h\times w \times c_1\times k\times k + h\times w\times c_1\times c_2,
\end{aligned}
\end{equation}
which is approximately $1/k^2$ compared to that of the corresponding standard convolutional layer. For instance, MobileNets~\cite{ref13,ref16} adopt $3\times3$ depthwise separable convolutions and have $8\times$ to $9\times$ less computational cost than the counterparts using standard convolutions. ShuffleNets~\cite{ref14,ref15} further utilize pointwise group convolutions, coupled with channel shuffling operations, to reduce the complexity of standard $1 \times 1$ pointwise convolutions.

\subsection{HBO Structure}

\begin{figure*}
	\centering
	\vspace{-2.5em}
	\includegraphics[width=0.9\linewidth]{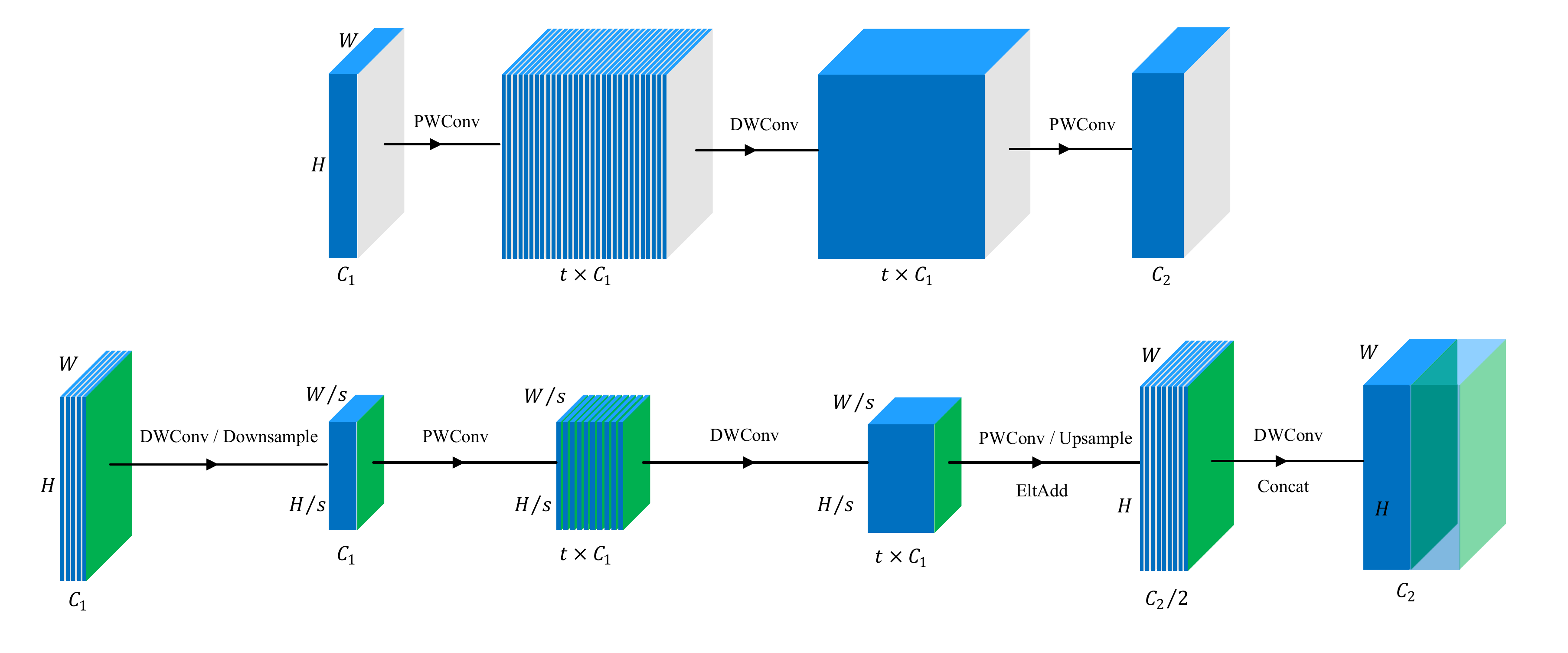}
	\vspace{-1.0em}
	\caption{
		Bottleneck comparison between MobileNetV2 and HBONet. The lower design represents our proposed HBO module while the upper one represents Inverted Residual with Linear Bottleneck in MobileNetV2. EltAdd means element-wise addition with corresponding channels in the input feature map, resembling the residual path. Concat means concatenation with partial channels of the input tensor. DWConv and PWConv denote depthwise and pointwise convolution respectively. Best viewed in color.
	}
	\label{fig:schema}
	\vspace{-1.0em}
\end{figure*}

By exploring the interdependencies among the channels of convolutional feature maps via a two-stage decomposition, depthwise separable convolution demonstrates impressive computation performance without noticeable sacrifice of accuracy compared to standard convolution. The characteristics of depthwise separable convolution render it especially fit for modern lightweight convolutional neural networks. In this line of research, how to design more powerful and efficient building blocks based on depthwise separable convolutions is the primary issue. Delving into recent state-of-the-art MobileNets~\cite{ref13,ref16} and ShuffleNets~\cite{ref14,ref15}, we notice that although various complex modular designs focusing on channel transformations have been invented to boost the performance within the limit of complexity, the spatial feature scale keeps the same across all compositional layers of these networks. This means the spatial feature dimension which is naturally complementary to the feature channel dimension in terms of accuracy and efficiency trade-off has never been explored. Hence we conjecture there is still exists remaining room to strike an improved balance between the representation capability and computational efficiency via further taking the spatial transformation into consideration, from a perspective orthogonal to aforementioned seminal works.

To this end, we introduce a novel bottleneck design, Harmonious Bottleneck on two Orthogonal dimensions (HBO), which consists of two reciprocal components, spatial contraction-expansion and channel expansion-contraction, nested in a bilaterally symmetric structure as illustrated in Figure~\ref{fig:schema} and functioning in a harmonious manner. Inverted residual blocks in MobileNetV2 reverse the classical configuration of bottlenecks for improved information flow. Nevertheless, the channel expansion-contraction transformation yields very wide feature maps in the middle of the building block, inevitably increasing the computational burden of relevant layers. We alleviate this problem by squeezing the channel expansion-contraction component, \ie, inverted bottleneck (in the channel dimension), into a pair of inverse spatial transformations which constitutes the spatial contraction-expansion component. This set of transformations from two orthogonal dimensions guarantees a slimmed spatial size of wide feature maps in each stage, mitigating a soaring consumption of computational resource arising from channel expansion operations. Benefiting from the inverse variation tendency of feature map size in two dimensions (spatial and channel), our proposed module tends to demand less computational resource against its straightforward counterparts given expected accuracy and is capable of retaining decent performance given limited computational cost. Subsequent layers responsible for upsampling are indispensable in most modular cases, which expand the spatial size of narrow feature maps for the convenience of spatial contraction operation in the follow-up HBO module, guaranteeing the depth of HBONet with hierarchical stacked HBO modules.

In the spatial contraction-expansion component, spatial contraction operation exploits the depthwise convolution with stride $s$ to downsample the spatial size of the input feature tensor from $h\times w\times c_1$ into $h/s\times w/s\times c_1$, and spatial expansion operation aims to upsample output features to make them have the identical spatial size with that of the input feature tensor, or probably its pooled version. After merging the spatial contraction-expansion component into existing blocks, the overall computational cost becomes
\begin{equation}\label{eq:02}
\begin{aligned}
B/{s^2}+(h/s\times w/s\times c_1+h\times w\times c_2)\times k^2,
\end{aligned}
\end{equation}
where $B$ denotes the original computational cost of the blocks inserted between the spatial contraction and expansion operations. Spatial contraction-expansion component,  which is also ready to be integrated into building blocks of any other state-of-the-art CNNs, demonstrates impressive flexibility and scalability.

\subsection{HBO Exemplars}
\begin{figure*}
	\centering
	\vspace{-2.0em}
	\includegraphics[width=0.7\linewidth]{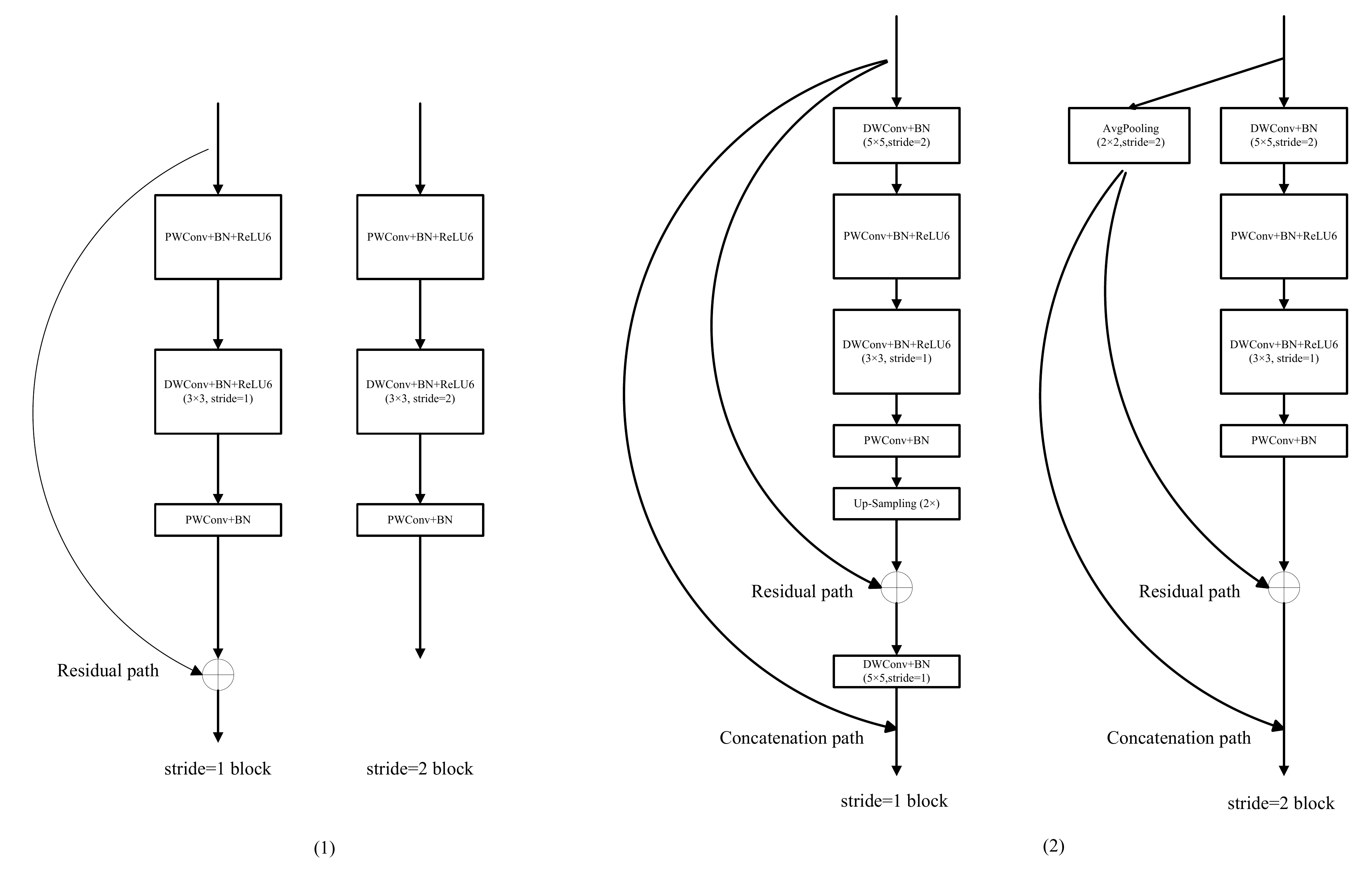}
	\caption{
		Basic modules of Inverted Residual with Linear Bottleneck and Harmonious Bottleneck, shown as (1) and (2) respectively. Each module is depicted under two circumstances with different settings of stride. Best viewed zoomed in.
	}
	\label{fig:module}
	\vspace{-1.5em}
\end{figure*}

As illustrated in Figure~\ref{fig:module}, we follow the design principle of layer modules in MobileNetV2 and use its body as our channel expansion-contraction pattern, where the low-dimensional representation is expanded in the channel dimension and filtered with an efficient depthwise convolution, subsequently contracted back to the space of low dimension with a linear convolutional filter. We go one step further to investigate transformations in the spatial dimension, through attaching a preceding depthwise convolution with stride and an optional subsequent bilinear up-sampling operation and its corresponding depthwise convolutional layer to the channel expansion-contraction module. We also follow the convention of including a residual path in the family of modern lightweight network architectures to facilitate the gradient propagation across multiple layers. Last but not least, half of the channels in the output feature map are drawn from the input tensor or its pooled version. This concatenation operation decreases the number of output channels to be computed in the main branch and encourages feature reuse in the information flow as an efficient and effective component.

\begin{table}[t]
	\centering
	\vspace{0pt}
	\resizebox{\linewidth}{!}{
		\begin{tabular}{c|c|c|c|c|c}
			\toprule[0.2em]
			Input size & Operator & $t$ & $c$ & $n$ & $s$ \\
			\toprule[0.2em]
			$224^2\times3$ & conv2d 3x3 & - & 32 & 1 & 2\\
			$112^2\times32$& Harmonious Bottleneck & 1 & 20 & 1 & 1\\
			$112^2\times20$& Harmonious Bottleneck & 2 & 36 & 1 & 1\\
			$112^2\times36$ & Harmonious Bottleneck & 2 & 72 & 3 & 2\\
			$56^2\times72$ & Harmonious Bottleneck & 2 & 96 & 4 & 2\\
			$28^2\times96$ & Harmonious Bottleneck & 2 & 192 & 4 & 2\\
			$14^2\times192$ & Harmonious Bottleneck & 2 & 288 & 1 & 1\\
			$14^2\times288$ & conv2d 1x1 & - & 144 & 1 & 1\\
			$14^2\times144$ & Inverted Residual & 6 & 200 & 2 & 2 \\
			$7^2\times200$ & Inverted Residual & 6 & 400 & 1 & 1 \\
			$7^2\times 400$ & conv2d 1x1 & - & 1600 & 1 & 1\\
			$7^2\times1600$ & avgpool 7x7 & - & - & 1 & - \\
			$1^2\times1600$ & conv2d 1x1 & - & k & - & \\
			\toprule[0.2em]
		\end{tabular}
	}
	\caption {
		\mbox{HBONet} : Each line describes a sequence of 1 or more identical (modulo stride)  layers, repeated $n$ times.
		All layers in the same sequence have the same number $c$ of output channels.
		The first layer of each sequence has a stride $s$ and all others use stride $1$.
		The expansion factor $t$ is always applied to the input size as described in Figure~\ref{fig:schema}.
	}
	\vspace{-1.5em}
	\label{hbonet:arch}
\end{table}

\subsection{Network Architecture}
Taking MobileNetV2~\cite{ref16} as a reference, we construct HBONet by stacking a set of HBO blocks and using them to replace some of the original blocks. We describe the architecture of HBONet in Table~\ref{hbonet:arch}, where Harmonious Bottleneck denotes our proposed building block while Inverted Residual denotes the reserved architecture unit as in MobileNetV2. Some other modifications are also made instead of performing a trivial replacement. For instance, we adjust the width with respect to each layer to approach a better balance between the model capacity and computational complexity. The expansion factor t in the micro-architecture is decreased compared with the original configurations of MobileNetV2, accommodating increased output channels c in the macro-architecture with the same level of computational budget. There also exists a pointwise convolution without subsequent nonlinear activation operation inserted between the two block groups of different types, projecting intermediate features into a low-dimensional representation space. Following a similar design principle as in MobileNetV2, we also provide a spectrum of models at different computational complexities.

%------------------------------------------------------------------------
\section{Experiments}
We conduct extensive experiments on several challenging benchmarks of visual recognition including image classification, object detection and person re-identification. Experimental results empirically demonstrate the scalability and efficiency of our proposed HBONet which is ready to be deployed in resource-aware platforms. All network architectures are constructed with the PyTorch \cite{paszke2017automatic} framework.

\subsection{Image Classification}
Our main experiments are performed to train the networks for the ImageNet~\cite{ref02} classification task. So far, both in academia and industry, ImageNet is known as the most famous image classification benchmark. It has about 1.2 million training images and 50 thousand validation images. The images in the dataset are natural images, and each image is annotated as one of 1000 object classes. We use scale and aspect ratio augmentation together with horizontal flipping as in \cite{ref06,ref08} to pre-process the dataset before feeding it into the networks for training. During evaluation, we follow a rescaling scheme that matches the smaller edges of images to the scale proportional to the training input size (\ie, divided by 0.875) and keeps their aspect ratios. Center regions of training input size are cropped from the resized images for single crop testing following common practice.

We re-implement the MobileNetV2 \cite{ref16} with a spectrum of width multipliers by ourselves. We keep the detailed optimization hyper-parameters totally the same for fair comparison when training our proposed corresponding networks. All models are trained with Stochastic Gradient Descend (SGD) with momentum for 150 epochs using batch size 256. The momentum is set as 0.9 and weight decay as 4e-5. The learning rate initiates from 0.05 and declines following a cosine function shaped decay strategy approximating to zero.

\begin{table}[htbp]
	\centering
	\resizebox{\linewidth}{!}{
		\begin{tabular}{l|ccc}
			\toprule [0.2em]
			Width Multiplier & Top-1 / Top-5 Acc. (\%) & MFLOPs & Top-1 Gain  \\
			\toprule [0.05em]
			MobileNetV2 (1.0) & 72.2 / 90.5 & 300  & --  \\
			HBONet (1.0) & \bf 73.1 / 91.0 & 305 & 0.9 \\
			\toprule [0.05em]
			MobileNetV2 (0.75) & 70.0 / 89.0 & 209 & -- \\
			HBONet (0.8) & \bf 71.3 / 89.7 & 205 & 1.3 \\
			\toprule[0.05em]
			MobileNetV2 (0.5) & 64.6 / 85.4 & 97 & -- \\
			HBONet (0.5) & \bf 67.0 / 86.9 & 96 & 2.4 \\
			\toprule[0.05em]
			MobileNetV2 (0.35) &  59.7 / 81.7 & 59 & -- \\
			HBONet (0.35) & \bf 62.4 / 83.7 & 61 & 2.7 \\
			\toprule[0.05em]
			MobileNetV2 (0.25) & 52.3 / 75.9 & 37 & -- \\
			HBONet (0.25) & \bf 57.3 / 79.8 & 37 & 5.0 \\
			\toprule[0.05em]
			MobileNetV2 (0.1) & 34.9 / 56.6 & 13 & -- \\
			HBONet (0.1) & \bf 41.5 / 65.7 & 14 & 6.6 \\
			\toprule [0.2em]
		\end{tabular}
	}
	\caption{
		Performance comparison on ImageNet validation set for different networks with varied width multipliers.
	}
	\label{table:width-comparison}
	\vspace{-2.0em}
\end{table}

Experimental results with networks in a spectrum of width multipliers are summarized in Table~\ref{table:width-comparison}, where the reported results of MobileNets are reproduced by ourselves which are comparative or higher than officially released results \cite{ref16}. For comparable complexity, we make minor adjustments to the configuration with respect to numbers of channels which are by default expected to be divisible by 8. Specifically, numbers of channels in MobileNetV2 (0.1) and our proposed counterpart are set to be divisible by 4, HBONets with width multipliers 0.5 and 0.25 set to be divisible by 2. From Table~\ref{table:width-comparison}, we observe that with the spatial contraction-expansion and channel expansion-contraction modules working collaboratively, our HBONet outperforms vanilla MobileNetV2 consistently under each level of complexity. Intriguingly, along with the decreasing complexity, gain of our HBONet over MobileNetV2 with the same level of computational cost tends to go larger. Especially under the computational budget of less than 40 MFLOPs, our network architectures still maintain decent performance, achieving impressive improvement upon MobileNetV2 which neglects transformation on the spatial dimension in its building blocks.

\begin{table}[htbp]
	\centering
	\resizebox{\linewidth}{!}{
		\begin{tabular}{l|ccc}
			\toprule [0.2em]
			Input Resolution & Top-1 / Top-5 Acc. (\%) & MFLOPs & Top-1 Gain  \\
			\toprule [0.05em]
			MobileNetV2 ($224 \times 224$) & 69.8 / 89.6 & 209  & --  \\
			HBONet ($224 \times 224$) & \bf 71.3 / 89.7 & 205 & 1.5 \\
			\toprule [0.05em]
			MobileNetV2 ($192 \times 192$) & 68.7 / 88.9 & 153 & -- \\
			HBONet ($192 \times 192$) & \bf 70.0 / 89.2 & 150 & 1.3 \\
			\toprule[0.05em]
			MobileNetV2 ($160 \times 160$) & 66.4 / 87.3 & 107 & -- \\
			HBONet ($160 \times 160$) & \bf 68.3 / 87.8 & 105 & 1.9 \\
			\toprule[0.05em]
			MobileNetV2 ($128 \times 128$) & 63.2 / 85.3 & 69 & -- \\
			HBONet ($128 \times 128$) & \bf 65.5 / 85.9 & 68 & 2.3 \\
			\toprule[0.05em]
			MobileNetV2 ($96 \times 96$) & 58.8 / 81.6 & 39 & -- \\
			HBONet ($96 \times 96$) & \bf 61.4 / 83.0 & 39 & 2.6 \\
			\toprule [0.2em]
		\end{tabular}
	}
	\caption{
		Performance comparison on ImageNet validation set for the same networks with varied input image resolutions. MobileNetV2 (0.75) and HBONet (0.8) with almost the same computational cost when the size of images fed into the networks is identical are selected for this comparison.
	}
	\label{table:resolution-comparison-0.75}
	\vspace{-1.0em}
\end{table}

\begin{table}[htbp]
	\centering
	\resizebox{\linewidth}{!}{
		\begin{tabular}{l|ccc}
			\toprule [0.2em]
			Input Resolution & Top-1 / Top-5 Acc. (\%) & MFLOPs & Top-1 Gain  \\
			\toprule [0.05em]
			MobileNetV2 ($224 \times 224$) & 60.3 / 82.9 & 59  & --  \\
			HBONet ($224 \times 224$) & \bf 62.4 / 83.7 & 61 & 2.1 \\
			\toprule [0.05em]
			MobileNetV2 ($192 \times 192$) & 58.2 / 81.2 & 43 & -- \\
			HBONet ($192 \times 192$) & \bf 60.9 / 82.6 & 45 & 2.7 \\
			\toprule[0.05em]
			MobileNetV2 ($160 \times 160$) & 55.7 / 79.1 & 30 & -- \\
			HBONet ($160 \times 160$) & \bf 58.6 / 80.7 & 31 & 2.9 \\
			\toprule[0.05em]
			MobileNetV2 ($128 \times 128$) & 50.8 / 75.0 & 20 & -- \\
			HBONet ($128 \times 128$) & \bf 55.2 / 78.0 & 21 & 4.4 \\
			\toprule[0.05em]
			MobileNetV2 ($96 \times 96$) & 45.5 / 70.4 & 11 & -- \\
			HBONet ($96 \times 96$) & \bf 50.3 / 73.8 & 12 & 4.8 \\
			\toprule [0.2em]
		\end{tabular}
	}
	\caption{
		Performance comparison on ImageNet validation set for the same networks with varied input image resolutions. MobileNetV2 (0.35) and HBONet (0.35) with almost the same computational cost when the size of images fed into the networks is identical are selected for this comparison.
	}
	\label{table:resolution-comparison-0.35}
	\vspace{-2.0em}
\end{table}

\begin{table}[htbp]
	\centering
	\vspace{-1.0em}
	\resizebox{\linewidth}{!}{
		\begin{tabular}{l|cc}
			\toprule [0.2em]
			Width \& Resolution & Top-1 / Top-5 Acc. (\%) & MFLOPs \\
			\toprule [0.05em]
			MobileNetV2 (0.5) ($224 \times 224$) & 64.6 / 85.4 & 97 \\
			HBONet (0.6) ($192 \times 192$) & \bf 67.7 / 87.4 & 98 \\
			\toprule [0.05em]
			MobileNetV2 (0.6) ($192 \times 192$) & 65.6 / 86.1 & 111 \\
			HBONet (0.5) ($224 \times 224$) & \bf 67.3 / 87.3 & 108 \\
			\toprule [0.2em]
		\end{tabular}
	}
	\caption{
		Performance comparison on ImageNet validation set for different networks with varied configurations of width multipliers and input image resolutions. Evaluations are performed after resizing the short sides of input images to 256, keeping their aspect ratios and cropping $224 \times 224$ regions in the center.
	}
	\label{table:cross-comparison}
	\vspace{-1.0em}
\end{table}

\begin{table}[htbp]
	\centering
	\vspace{0.5em}
	\resizebox{\linewidth}{!}{
		\begin{tabular}{c|cc}
			\toprule [0.2em]
			Down / Up-sampling rate & Top-1 / Top-5 Acc. (\%) & MFLOPs \\
			\toprule [0.05em]
			HBONet ($2\times$) & 58.3 / 80.6 & 44 \\
			HBONet ($4\times$) & {\bf 59.3 / 81.4} & 45\\
			HBONet ($8\times$) & 58.2 / 80.4 & 45 \\
			\toprule [0.2em]
		\end{tabular}
	}
	\caption{
		Performance comparison on ImageNet validation set for different networks with varied spatial contraction units. HBONet (0.25) is utilized as the test case. Notice that numbers of channels in the HBONet (0.25) here are set to be divisible by 8, which is different from the one in Table~\ref{table:width-comparison}. A variant network is denoted as HBONet ($2^k\times$) if the maximum number of spatial contraction units in one block is k as illustrated in Figure~\ref{fig:variant}.
	}
	\label{table:variant-comparison}
	\vspace{-1.0em}
\end{table}

We further conduct experiments on a spectrum of input resolutions for comparison. From Table~\ref{table:resolution-comparison-0.75} and Table~\ref{table:resolution-comparison-0.35}, it is evident that our proposed HBONet also outperforms vanilla MobileNetV2 with varied input image resolutions. It shows the similar trend that decreasing computational cost leads to larger margin between each group of rivals with the similar complexity. The performance of MobileNetV2 in Table~\ref{table:resolution-comparison-0.75} and Table~\ref{table:resolution-comparison-0.35} is collected from the official GitHub page of TensorFlow\footnote{\url{https://github.com/tensorflow/models/tree/master/research/slim/nets/mobilenet}}.

We also take into consideration the trade-off between width multiplier and input size, two group of networks with similar complexity but different width multiplier and input size are selected towards our verification goal. As demonstrated in Table~\ref{table:cross-comparison}, we find that our proposed networks achieve consistent improvement regardless of the combination of width multiplier and input resolution, demonstrating the superiority of our architecture over previous engineered blocks mainly comprising of plain depthwise separable convolutions.

\begin{figure}
	\includegraphics[width=\linewidth]{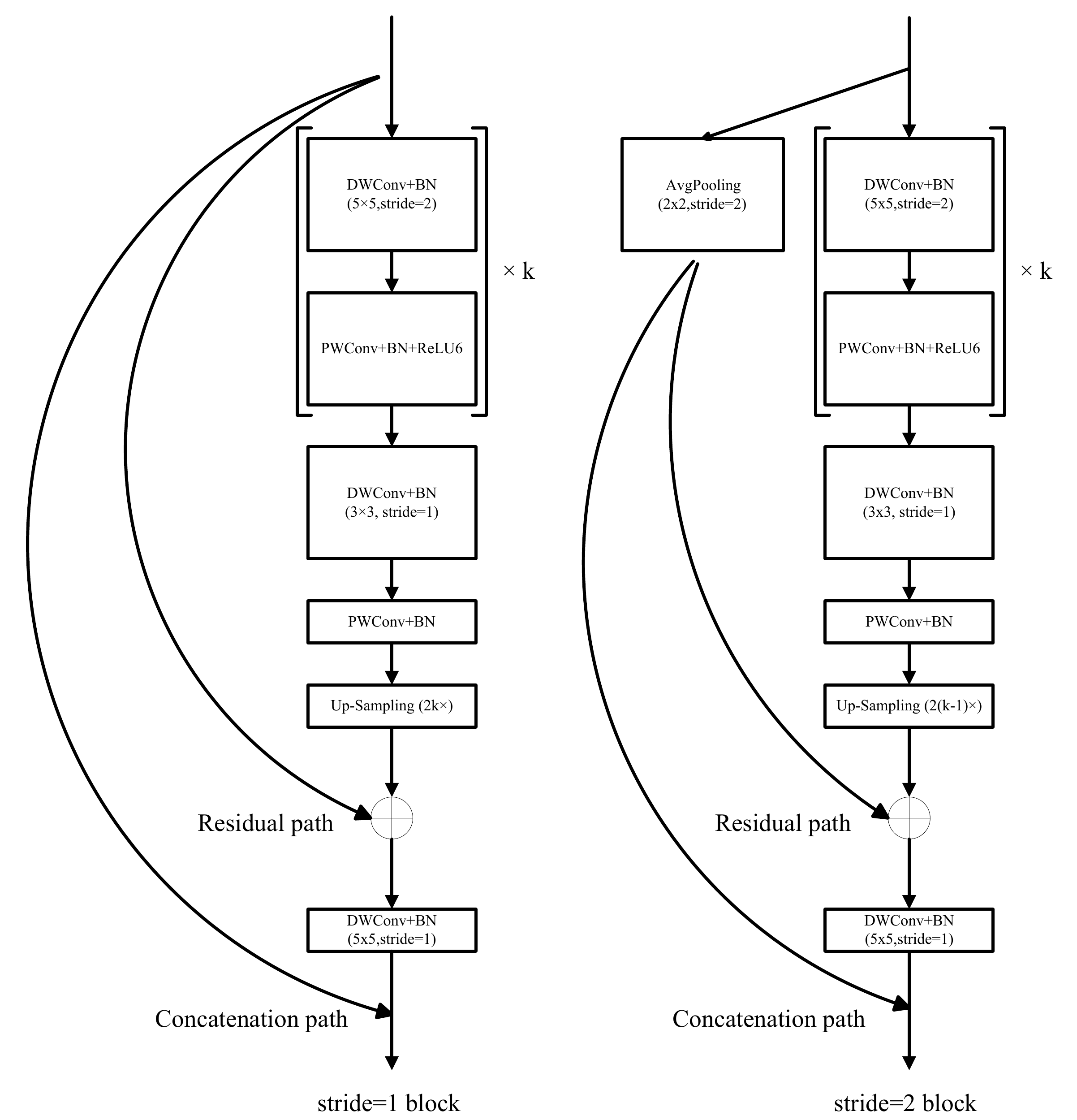}
	\caption{Schema of variant blocks including k successive spatial contraction units.}
	\label{fig:variant}
	\vspace{-1.0em}
\end{figure}

\textbf{Variants with cascade spatial contraction units.} For further exploration on the benefit of our novel spatial encoding methodology, we stack depthwise separable convolutions with stride of 2 at the front-end and use a larger up-sampling rate to restore the size of input feature maps or its pooled version at the back-end. See Figure~\ref{fig:variant} for detailed architecture of our variants. When more spatial contraction units are inserted in one block, we only preserve the non-linear activation operations at the start and the end respectively, guaranteeing the linearity of our proposed building block, and thus the network depth is not increased. With this extension, we achieve further improvement compared with our proposed basic modular design as illustrated in Table~\ref{table:variant-comparison}, opening up a direction to explore in the future work.

The statistics with respect to computational complexity and classification accuracy about a series of relevant compact models are summarized in Table~\ref{table:performance-comparison} for reference.

\begin{table}[htbp]
	\centering
	\resizebox{\linewidth}{!}{
		\begin{tabular}{l|cc}
			\toprule [0.2em]
			Networks & Top-1 Acc. (\%) & FLOPs \\
			\toprule [0.2em]
			MobileNetV1 (0.75) \cite{ref13} & 69.5 & 325M \\
			CondenseNet (G=C=8) \cite{ref22} & 71.0 & 274M  \\
			ShuffleNetV1 $1.5 \times$ (g=3) \cite{ref14} & 71.5 & 292M \\
			MobileNetV2 (1.0) \cite{ref16} & 72.0 &  300M \\
			IGCV3-D \cite{ref24} & 72.2 & 318M \\
			ShuffleNetV2 $1.5 \times$ \cite{ref15} & 72.6 & 299M \\
			HBONet (1.0) & {\color{blue} 73.1} & 305M  \\
			\toprule [0.2em]
		\end{tabular}
	}
	
	\caption{
		Performance comparison of several efficient networks over top-1 classification accuracy on \mbox{ImageNet}  validation set with single crop. As is common practice for FLOPs, we count the total number of Multiply-Adds. The top-1 accuracy of our proposed HBONet (1.0) is highlighted in {\color{blue} blue}, surpassing all the other state-of-the-art networks under the complexity level of around 300 MFLOPs.
	}
	\label{table:performance-comparison}
	\vspace{-1.5em}
\end{table}

\subsection{Object Detection}
We also evaluate and compare the generalization ability of our proposed HBONet and MobileNetV2 on the PASCAL VOC object detection benchmark \cite{Everingham2015}. We perform experiments with the fast single-stage detection framework, Single Shot Detector (SSD) \cite{10.1007/978-3-319-46448-0_2}, using backbones with varied width multipliers described in the previous sub-section as feature extractors. Our evaluation aims at comparing the efficiency of backbone networks thus we keep the detection heads for classification and localization the same when adjusting the width of backbones. The specific setup (\eg attached locations in the backbone, structure of convolutional layers, size of corresponding prior boxes, \etc) of these extra prediction layers are consistent with \mbox{MobileNetV2 + SSD} \cite{ref16} in all of our experiments.

\begin{table}[htbp]
	\centering
	\resizebox{.8\linewidth}{!}{
		\begin{tabular}{l|cc}
			\toprule [0.2em]
			Width Multiplier & mAP (\%) & Gain  \\
			\toprule [0.05em]
			MobileNetV2 SSD320 (1.0) & 70.4 & --  \\
			HBONet SSD320 (1.0) & \bf 71.0 & 0.6 \\
			\toprule [0.05em]
			MobileNetV2 SSD320 (0.5) & 63.6 & -- \\
			HBONet SSD320 (0.5) & \bf 64.8 & 1.2 \\
			\toprule[0.05em]
			MobileNetV2 SSD320 (0.25) & 51.6 & -- \\
			HBONet SSD320 (0.25) & \bf 55.9 & 4.3 \\
			\toprule[0.05em]
			MobileNetV2 SSD320 (0.1) & 36.3 & -- \\
			HBONet SSD320 (0.1) & \bf 42.6 & 6.3 \\
			\toprule[0.05em]
			\toprule[0.05em]
			FD-MobileNet SSDLite \cite{8451355} & 62.1 & -- \\
			MobileNet SSD300 \cite{Huang_2017_CVPR} & 68 & -- \\
			Pelee \cite{NIPS2018_7466} & 70.9 & -- \\
			\toprule [0.2em]
		\end{tabular}
	}
	\caption{
		Performance comparison on PASCAL VOC 2007 test set with SSD on $320 \times 320$ resolution for MobileNetV2 and HBONet. Different backbones with varied widths are evaluated for a more comprehensive benchmark.
	}
	\label{table:pascal}
	\vspace{-2.0em}
\end{table}

We train all the models on the union of PASCAL VOC 2007 \textit{trainval} set and 2012 \textit{trainval} set. Our training scheme primarily follows the original SSD \cite{10.1007/978-3-319-46448-0_2}, including data augmentation and hard example mining process and so forth. We set the batch size as 32 and train for 560 epochs in total. We adopt SGD with momentum as the default optimizer, with the momentum set as 0.9 and the weight decay as 5e-4. The initial learning rate of original SSD starts at 1e-3. For better convergence, we utilize the warm-up strategy which linearly ramps up the learning rate from a close-to-zero one (\ie, 1e-6) to the normal initial learning rate of 1e-3 during the first 5 epochs. When the learning rate goes back to the original schedule, it is divided by 10 at the epoch 360 and 480 respectively. Parallel to SSD with MobileNetV2, we resize the input image size to $320 \times 320$.

Evaluation results are reported under the protocol of mean Average Precision (mAP) on the PASCAL VOC 2007 \textit{test} set in Table~\ref{table:pascal}. We observe that similar to the main experiments on ImageNet dataset, SSD with narrower HBONets as backbones outperforms the corresponding \mbox{MobileNetV2 + SSD} to a greater extent. The comparison further demonstrates the improved representation ability of our proposed backbone network over MobileNetV2 on the more challenging object detection problem, especially in the extremely resource-constrained conditions. We also include the results of other detection frameworks using lightweight networks as feature extractors in Table~\ref{table:pascal} for reference.

\subsection{Person Re-Identification}
We finally perform experiments on the popular person re-identification dataset Market-1501 to address instance level recognition problems. The Market-1501 \cite{Zheng_2015_ICCV} dataset consists of 12,936 training images, 15,913 gallery images and 3368 queries, in which bounding boxes out of 1501 identities are captured by 6 cameras in front of the supermarket inside the campus of Tsinghua University.

\begin{table}[htbp]
	\centering
	\resizebox{.8\linewidth}{!}{
		\begin{tabular}{l|c|c|c|c}
			\toprule[0.2em]
			\multirow{2}*{Width Multiplier} & \multicolumn{2}{|c|}{Performance} & \multicolumn{2}{|c}{Gain} \\
			\cline{2-5}
			& mAP & Rank-1 & mAP & Rank-1 \\
			\toprule[0.2em]
			MobileNetV2 (1.0) & 70.5 & 88.5 & -- & -- \\
			HBONet (1.0) & \bf 74.4 & \bf 90.2 & 3.9 & 1.7 \\
			\toprule[0.05em]
			MobileNetV2 (0.5) & 67.3 & 86.6 & -- & -- \\
			HBONet (0.5) & \bf 71.0 & \bf 88.7 & 3.7 & 2.1 \\
			\toprule[0.05em]
			MobileNetV2 (0.25) & 60.1 & 81.7 & -- & -- \\
			HBONet (0.25) & \bf 63.7 & \bf 84.5 & 3.6 & 2.8 \\
			\toprule[0.05em]
			MobileNetV2 (0.1) & 43.7 & 68.3 & -- & -- \\
			HBONet (0.1) & \bf 47.7 & \bf 73.2 & 3.9 & 5.0 \\
			\toprule[0.05em]
			\toprule[0.05em]
			ResNet-50 & 70.3 & 88.5 & -- & -- \\
			\toprule[0.2em]
		\end{tabular}
	}
	\caption{Comparison of Rank-1 accuracy and mAP on the Market-1501 dataset with different networks with a spectrum of width multipliers as backbones.}
	\label{tab:market}
	\vspace{-2.0em}
\end{table}

We adopt our HBONets and their corresponding MobileNetV2 described above as the backbone networks. For training, image samples are rescaled slightly larger than the target size, then cropped to the target size of $256 \times 128$ randomly. Horizontal flipping and normalization are adopted as the common data augmentation techniques. The re-ID models are trained with the AMSGRAD \cite{j.2018on} optimizer ($\beta_1=0.9,\beta_2=0.999$, weight decay=5e-4) for 90 epochs using batch size 32. The learning rate initiates at 0.001 and is decayed by a factor of 0.1 at the epoch of 60. For the relatively large HBONet (1.0) and its MobileNetV2 counterpart, we fine-tune for another 30 epochs with decayed learning rate for better convergence. We also apply label smooth techniques during the whole training process since images in the re-ID datasets are not diverse enough. The adapted fully-connected classifier for re-ID models, without available pre-trained weights loaded, is trained for 10 epochs in advance. During this warm-up stage, all the other layers are fixed.

From Table~\ref{tab:market}, we observe that substantial and increasing gains are achieved using our HBONets with shrinking width multipliers, though there shows a saturating trend in terms of the mAP evaluation. The compelling gain in the instance recognition problem further demonstrates the power of our proposed backbone. Corresponding results using the prevalent ResNet-50 are also included in the last row for reference.

\section{Conclusion}
In this paper, we propose HBO, a compact bottleneck specially designed to improve the performance of the class of lightweight CNNs based on depthwise separable convolutions under extremely limited computational budget (\eg less than 40 MFLOPs). HBO jointly models the interdependencies across spatial and channel dimensions of depthwise convolutional features via a bilaterally symmetric structure consisting of a spatial contraction-expansion component and a channel expansion-contraction component. Extensive experiments on several datasets show the effectiveness of HBONets constructed with HBO modules.

{\small
\bibliographystyle{ieee_fullname}
\bibliography{egbib}
}

\end{document}